\documentclass[12pt]{article}
\usepackage{microtype}
\usepackage[title]{appendix}
\usepackage{graphicx}
\usepackage{subfigure}
\usepackage{amsmath}
\usepackage{booktabs} % for professional tables
\usepackage{longtable}
\usepackage{amsmath,amssymb, amsthm}
\usepackage[a4paper,margin=1.375in]{geometry}

\title{Complex Amplitude-Phase Boltzmann Machines}
\author{% 
\textbf{Zengyi Li}$^{2,3}$ \hspace{1cm} \textbf{Friedrich T. Sommer} $^{1,3,4}$\\
$^1$Intel Labs, Santa Clara, CA 95054-1549\\
$^2$Department of Physics \\
$^3$Redwood Center for Theoretical Neuroscience\\
$^4$Helen Wills Neuroscience Institute \\
University of California, Berkeley\\
Berkeley, CA 94720\\
\texttt{zengyi\_li@berkeley.edu, fsommer@berkeley.edu}
}

\date{January 2020}

\begin{document}

\maketitle

\begin{abstract}
  We extend the framework of Boltzmann machines to a network of complex-valued neurons with variable amplitudes, referred to as Complex Amplitude-Phase Boltzmann machine (CAP-BM). The model is capable of performing unsupervised learning on the amplitude and relative phase distribution in complex data. The sampling rule of the Gibbs distribution and the learning rules of the model are presented. Learning in a Complex Amplitude-Phase restricted Boltzmann machine (CAP-RBM) is demonstrated on synthetic complex-valued images, and handwritten MNIST digits transformed by a complex wavelet transform. Specifically, we show the necessity of a new amplitude-amplitude coupling term in our model. The proposed model is potentially valuable for machine learning tasks involving complex-valued data with amplitude variation, and for developing algorithms for novel computation hardware, such as coupled oscillators and neuromorphic hardware, on which Boltzmann sampling can be executed in the complex domain.
\end{abstract}

\section{Introduction}
Boltzmann machines are recurrent stochastic neural networks that can be used for learning data distributions. Originally proposed with binary stochastic neurons \cite{ackley1985BMlearnining}, a complex-valued Boltzmann machine was first introduced under the name DUBM (Directional Unit Boltzmann Machine) \cite{zemel1995DUBM}. In this model, the neurons represent complex numbers of modulus 1 with arbitrary phase angles. DUBM can learn relative phase distributions. The practical impact of DUBM has been somewhat limited because complex data representing real-world problems often have not only phase but also amplitude variations. From a neuroscience perspective, DUBMs also have the undesirable property that all neurons are active all the time.   
Here we propose a complex Boltzmann machine whose neurons can represent complex numbers with arbitrary phase angles and amplitudes of 1 or 0. As we demonstrate in simulation experiments, this model enables unsupervised learning of complex-valued data with variable amplitudes. Further, it permits the introduction of regularization of the network activity, such as a sparsity constraint. We also show the necessity of an amplitude-amplitude coupling term that is potentially useful for other types of complex-valued neural networks \cite{trabelsi2018deepcomplexnet}\cite{guberman2016complexCNN}.

\section{Model Setup}
The DUBM model \cite{zemel1995DUBM} is an energy based model, $p(\pmb{z}) = e^{-E(\pmb{z})}/Z$, for a data distribution of phasor variables, i.e., a vector of complex-valued components $z_{j}$ with modulus 1. $Z$ is the partition sum. The energy function of the DUBM is given by:
\begin{equation}
    E(\pmb{z}) = -\frac{1}{2}\pmb{z}^{\dagger}\pmb{Wz}
    \label{eq_energy}
\end{equation}
where the superscript $\dagger$ denotes the conjugate transpose. The matrix $\pmb{W} \in \mathbb{C}^N$ is a complex coupling matrix. For (\ref{eq_energy}) to be real-valued, the matrix is required to be Hermitian, i.e.,  $\pmb{W}^\dagger = \pmb{W}$. 

If we allow a state $z_j$ to take two modulus values, $1$ and $0$, corresponding to an active or inactive neuron, (\ref{eq_energy}) induces an amplitude and relative phase distribution. To control the fraction of active units, we add into (\ref{eq_energy}) a penalty term of the form: $\pmb{\epsilon}^{T}|\pmb{z}|$, where $\pmb{\epsilon} \in \mathbb{R}^N$ is a bias vector. Further, we introduce an amplitude-amplitude coupling term: $-\frac{1}{2}|\pmb{z}|^T\pmb{J}|\pmb{z}|$, with $\pmb{J} \in \mathbb{R}^N$ a symmetric real-valued matrix. Putting it all together, the energy function of Complex Amplitude-Phase Boltzmann Machine (CAP-BM) is:
\begin{equation}
    E(\pmb{z}) = -\frac{1}{2}\pmb{z}^{\dagger}\pmb{Wz} -\frac{1}{2}|\pmb{z}|^T\pmb{J}|\pmb{z}|  +\pmb{\epsilon}^{T}|\pmb{z}|
\label{apbm_energy}
\end{equation}
Like in the DUBM model, the CAP-BM model is symmetric with respect to global phase shifts in all units. The benefit of the amplitude-amplitude coupling in the CAP-BM might not be obvious here, but we will explore its effect experimentally and argue later why this term is essential.  

Sampling in complex Boltzmann machines can be achieved by a Gibbs sampling procedure similar to that in real-valued Boltzmann machines. One difference is that we sample amplitude and phase separately. To achieve this, two marginal probabilities induced by the Boltzmann distribution are required: $P(|z_j|=1|\,\pmb{z}_{!j})$ and $p(\theta_{j}|\,|z_j|=1,\pmb{z}_{!j})$. They represent the marginal probability for a unit to take amplitude $1$ and the probability density of its phase, given that it takes amplitude $1$. They can be obtained in the same manner as in \cite{ackley1985BMlearnining}, for derivations, see Appendix A:
\begin{eqnarray}
P(|z_j|=1|\,\pmb{z}_{!j}) &=& \frac{1}{1+(e^{\mu_j-\epsilon_j} \mathrm{I_0} (a_j))^{-1}}
\label{pjayfinal}\\
p(\theta_{j}|\,|z_j|=1,\pmb{z}_{!j})
&=&\frac{1}{\mathrm{2\pi I_0}(a_j)}e^{a_j \mathrm{cos}(\alpha_j-\theta_j)}
\label{ptheta}
\end{eqnarray}
In the above equations the variables $a_j, \alpha_j, \mu_j$ represent the complex and real-valued input sums to neuron $j$: $u_j = a_j e^{i\alpha_j}= \sum_{k\neq j} W_{jk}z_k$ and $\mu_j = \sum_{k\neq j} J_{jk}|z_k|$. 
$\mathrm{I_0}(x)$ denotes the zeroth order Bessel function of the first kind, which becomes similar to an exponential function for large arguments. Therefore, $P(|z_j|=1|\,\pmb{z}_{!j})$ is sigmoid shaped as a function of $a_j$ and $\mu_j$. Similar as that in the DUBM model, the phase distribution $p(\theta_{j}|\,|z_j|=1,\pmb{z}_{!j})$ is a von Mises distribution, the circular analog of Gaussian. For a graphic depicting of the behavior of $P(|z_j|=1|\,\pmb{z}_{!j})$, see Appendix, Figure~\ref{fig:appendix}.

Note here the amplitude depends on phase through $a_j$, and phase depends on amplitude as units with amplitude $0$ do not contribute to the input sum $u_j$. Therefore, the CAP-BM model is not equivalent to the combination of DUBM and a real-valued Boltzmann machine, in which amplitudes and phases would be modeled separately. 

\section{Learning rules for the Complex Boltzmann machine}
Like for the real-valued Boltzmann machine\cite{ackley1985BMlearnining}, the learning rules for model parameters of the CAP-BM model can be derived for the Maximum Likelihood objective $G$ (derivations, see Appendix):
\small
\begin{eqnarray}
%\begin{split}
\frac{\partial G}{\partial b_{jk}} 
%= \left<\left<\frac{\partial E(\pmb{v}, \pmb{h})}{\partial b_{jk}}\right>_{p(\pmb{h}|\,\pmb{v})}-\left< \frac{\partial E(\pmb{v}', \pmb{h}')}{\partial b_{jk}}\right>_{p(\pmb{v}',\,\pmb{h}')}\right>_{p(\pmb{v})} \\
&=&\left<|z_j||z_k|\mathrm{cos}(\theta_{jk}+\theta_k-\theta_j)\right>_{\mathrm{sample}}-\left<|z_j||z_k|\mathrm{cos}(\theta_{jk}+\theta_k-\theta_j)\right>_{\mathrm{model}}\label{eq_rule1}\\
%\end{split}
%\end{equation}
%\normalsize
%Similarly, the gradients w.r.t $\theta_{jk}$, $J_{jk}$ and $\epsilon_j$ are:
%\small
%\begin{equation}
\frac{\partial G}{\partial \theta_{jk}}
&=& \!\!\!\!\!-\!\left<|z_j||z_k|b_{jk}\mathrm{sin}(\theta_{jk}\!+\!\theta_k\!-\!\theta_j)\right>_{\mathrm{sample}} \!\!+\! \left<|z_j||z_k|b_{jk}\mathrm{sin}(\theta_{jk}\!+\!\theta_k\!-\!\theta_j)\right>_{\mathrm{model}}\label{eq_rule2}\\
%\end{equation}
%\begin{equation}
\frac{\partial G}{\partial J_{jk}}
&=& \left<|z_j||z_k|\right>_{\mathrm{sample}}-\left<|z_j||z_k|\right>_{\mathrm{model}}\label{eq_rule3}\\
\frac{\partial G}{\partial \epsilon_{j}}
&=& -\left<|z_j|\right>_{\mathrm{sample}}+\left<|z_j|\right>_{\mathrm{model}}\label{eq_rule4}
\end{eqnarray}
\normalsize
Here $b_{jk}$ and $\theta_{jk}$ denote amplitude and phase of complex weight $W_{jk}=b_{jk}e^{i \theta_{jk}}$, $J_{jk}$ the real-valued weight for amplitude-amplitude coupling, and $\epsilon_{j}$ the bias. 

The learning rules (\ref{eq_rule3}) and (\ref{eq_rule4}) are the same as that of real-valued BM while rules (\ref{eq_rule1}) and (\ref{eq_rule2}) are similar to that of DUBM with extra amplitude dependencies. Another similarity to real-valued BMs is that training in our model requires sampling from the model distribution. To speed up the training in real-valued BMs, learning schemes such as Contrastive Divergence (CD) and Persistent Contrastive Divergence (PCD) \cite{hinton2002CD,tieleman2008PCD} have been proposed that do not require full model distribution. Another proposal for higher sampling efficiency is to choose a network architecture, now called the restricted Boltzmann machine (RBM) \cite{smolensky1986information}, in which sampling from model is more parallelizable because recurrent weights within the sets of hidden or visible units are absent. All these techniques for speeding up the training can equally be applied to the CAP-BM model.

\section{Experiments with a Complex Phase-Amplitude RBM}

%The proposed framework can be applied to the special case of a Restricted Boltzmann Machine in the obvious way. 
Here we demonstrate a restricted version of the CAP-BM, referred to as CAP-RBM, on synthetic data and on the MNIST dataset pre-processed with a complex wavelet transform (CWT) (for details, see Appendix.~\ref{Appendix: exp_detail}).
\begin{figure}[H]
    \centering
    \includegraphics[scale=1]{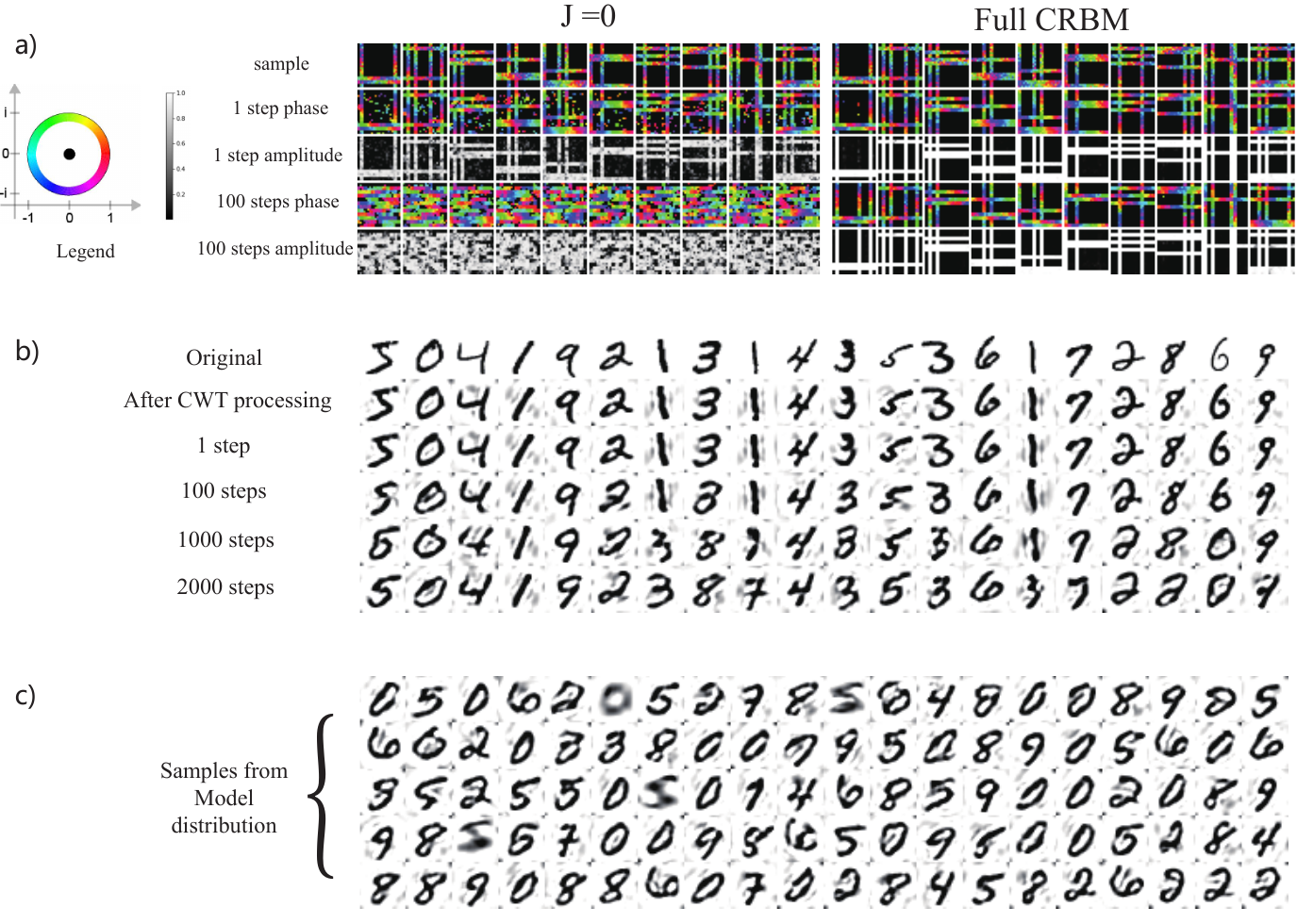}
    % \vspace{-0.10in}
    \caption{Demonstration of the Complex restricted Boltzmann machine (CAP-RBM) on a synthetic bar dataset and on CWT-transformed MNIST dataset. a) Training CAP-RBM on complex images of bars with noisy phase (best viewed in color). The two blocks of images show results of the CAP-RBM without $J$ matrix and the full CAP-RBM. The first row in each block shows samples from dataset. The four lower rows in each block show the expectation of visible unit activity after variable numbers of Gibbs sampling steps initialized at sample. The model without $J$ matrix does not form stable representation of the sample. b) original and reconstructed MNIST digits after various numbers of sampling steps, initialized at samples. c) samples generated from random initialization, global phase for each sample has to be set by hand.}
    
    \label{fig:main}
\end{figure}

For synthetic dataset, the CAP-RBM was trained using 1-step contrastive divergence (CD-1) \cite{hinton2002CD} on a synthetic dataset of complex-valued images of bars with a noisy sine-wave phase pattern. We compare the performance of models with and without the amplitude-amplitude coupling term $J$. As can be seen in Fig.~\ref{fig:main} a), the model without the $J$ term does not form a stable representation of test data.

The necessity of $J$ term can be explained as follows. In equations~\ref{ptheta} and \ref{pjayfinal} one can see that the amplitude of the complex input sum to a unit, $a_j$, plays a dual role of controlling the activation of a unit and the variance of phase distribution. Sometimes the data may have sharp amplitude distribution while having large variance on its phase, this distribution cannot be learned without $J$ since this would require $a_j$ to be large and small at the same time.

We then train CAP-RBM on complex wavelet transformed MNIST dataset, where only middle two frequency bands are used and the complex coefficients are thresholded and normalized. Training used PCD \cite{tieleman2008PCD} algorithm after initializing with CD-1. As can be seen from Fig.~\ref{fig:main} b) and c), the model captures data distribution well.
%, where only middle two frequency bands are used and the complex coefficients are thresholded and normalized before used for training, see Appendix for Fortunately the reconstruction is not strongly affected as the structure of the image is largely contained in the relative phase patterns. 

\section{Conclusion}

%Discuss \cite{nakashika2017complex}.

In this report we proposed and demonstrated a model of Boltzmann machine with both amplitude and phase variation. Our work differs from previous formulations of complex BM by being a natural extension of DUBM. In contrast, \cite{popa2018complexbinaryDBM} use variables with binary real and imaginary parts, and \cite{nakashika2017complex} use complex-Gaussian visible unit. In particular we showed the importance of an amplitude-amplitude coupling term not seen in previous works on complex-valued neural networks. In addition, this model is potentially directly applicable since new hardware implementation of Boltzmann sampling in complex domain is becoming available. Examples include electronic \cite{wang2019oscillatorIsingmachinereport} and  optical\cite{takeda2017boltzmannsamplingbyoptics} implementations. Furthermore, there is recent proposal of mapping recurrent network of oscillating spiking neurons to complex networks \cite{frady2019TPAM}, which could also benefit from a probabilistic interpretation. 

Acknowledgements: ZL has been supported by a research gift of the Intel Neuromorphic Research Community, FTS has been partly supported by grant 1R01EB026955-01 from the National Institute of Health.

\bibliographystyle{apa}
\bibliography{reference}

\newpage
\begin{appendices}
\section{Sampling rules derivation}
\begin{figure}[!th]
    \centering
    \includegraphics[scale=1]{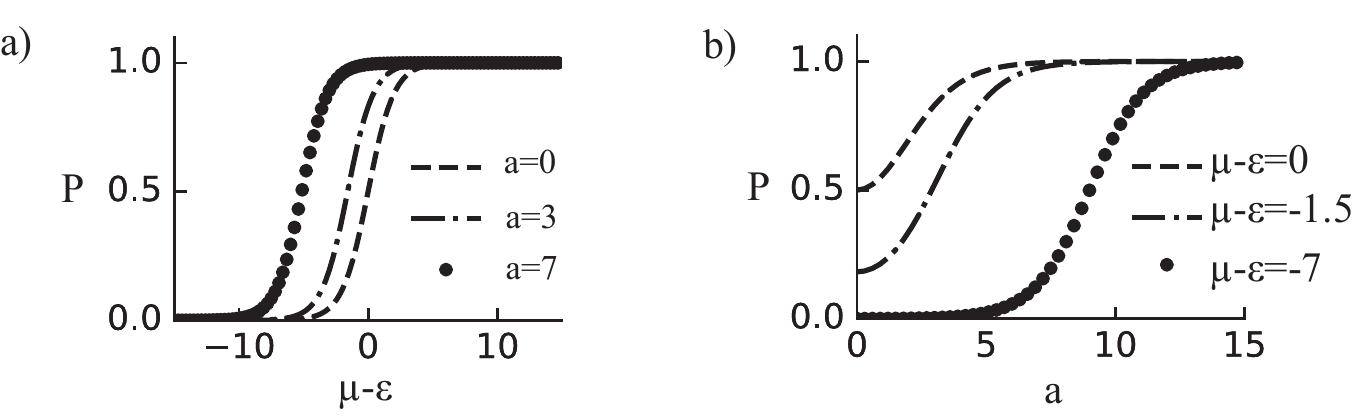}
    % \vspace{-0.10in}
    \caption{The amplitude activation function $P_j$. a) $P_j$ is a sigmoid function with respect to the real-valued postsynaptic sum with the saturation levels at $P_j=0$ and $P_j=1$. A horizontal offset is determined by $a$. b) T$P_j$ as a function of the modulus of the complex postsynaptic sum is also approximately sigmoid shaped except near $a=0$ where it always have slop $0$. For large negative values of $\mu - \epsilon$, saturation levels are at $P_j=0$ and $P_j=1$. However, when $\mu - \epsilon = 0$, the value of $P_j$ near $a=0$ rises to a value of 0.5.}
    
    \label{fig:appendix}
\end{figure}

The amplitude probability distribution $P(|z_j|=1|\,\pmb{z}_{!j}) =:P_j$ is computed as marginals of the Boltzmann distribution induced by energy function (\ref{apbm_energy}): 
%We find $P_j$ first:
\begin{equation}
P_j =\int_{2\pi} d\theta \, p(|z_j|=1,\theta_j = \theta| \, \pmb{z}_{!j}) 
=\frac{\int_{2\pi} d\theta \, e^{-E_j(|z_j|=1,\theta_j = \theta)-E_{!j}}}{\pmb{Z}}
\label{pjay}
\end{equation}
Here $E_j$ and $E_{!j}$ denote parts of the energy function that depend and do not depend on $z_j$, respectively. To avoid dealing with the intractable partition function $\pmb{Z}$, one can calculate $P_j/1-P_j$ then solve for $P_j$:
\begin{equation}
1-P_j = \frac{\int_{2\pi} d\theta \, e^{-E_j(|z_j|=0,\theta_j = \theta)-E_{!j}}}{\pmb{Z}}
\label{oneminuspjay}
\end{equation}
Divide (\ref{pjay}) and (\ref{oneminuspjay}) and insert the following expression easily derived from (\ref{apbm_energy}): $E_j = -Re(z_j^\ast u_j) - |z_j|\mu_j +\epsilon_j$ where 
$u_j = \sum_{k\neq j} w_{jk}z_k = a_j e^{i\alpha_j}$ and 
$\mu_j = \sum_{k\neq j} J_{jk}|z_k|$ is the complex and real-valued postsynaptic sums, respectively.
Putting it all together yields:
\small
\begin{equation}\nonumber
\frac{P_j}{1-P_j} = \frac{\int_{2\pi} d\theta \, e^{-E_j(|z_j|=1,\theta_j = \theta)-E_{!j}}}{\int_{2\pi} d\theta \, e^{-E_{!j}}}
= \frac{1}{2\pi}\int_{2\pi}d\theta \, e^{a_j \mathrm{cos}(\alpha_j-\theta) +\mu_j -\epsilon_j}
= e^{\mu_j-\epsilon_j}\mathrm{I_0} (a_j)
\end{equation}
\normalsize
where $\mathrm{I_0}$ denotes the zeroth order modified Bessel function of the first kind. Solving for $P_j$ yields:
\begin{equation}
%P_j(a_j,\mu_j,\epsilon_j) 
P(|z_j|=1|\,\pmb{z}_{!j}) = \frac{1}{1+(e^{\mu_j-\epsilon_j} \mathrm{I_0} (a_j))^{-1}}
\label{pjayfinal}
\end{equation}
We note that result (\ref{pjayfinal}) can also serve as the natural amplitude activation function for a continuous-valued complex neural network that also has amplitude-amplitude coupling. See Figure~\ref{fig:appendix} for a plot illustrating some of its properties.

Finally, for obtaining the phase distribution $p(\theta_{j}|\,|z_j|=1,\pmb{z}_{!j})$, we use Bayes rule: 
\small
\begin{equation}
\begin{split}
p(\theta_{j}|\,|z_j|=1,\pmb{z}_{!j})
=\frac{p(|z_j|=1,\theta_j \, |\pmb{z}_{!j})}{P(|z_j|=1\,|\pmb{z}_{!j})} 
& =\frac{e^{-E_j(|z_j|=1,\theta_j)-E_{!j}}}{\int_{2\pi} d\theta \, e^{-E_j(|z_j|=1,\theta_j = \theta)-E_{!j}}} \\
& =\frac{1}{2\pi \mathrm{I_0}(a_j)}e^{a_j \mathrm{cos}(\alpha_j-\theta_j)}
\end{split}
\end{equation}
\normalsize
which is a von Mises distribution, the circular analog of Gaussian with its mode at $\theta_j=\alpha_j$.

\section{Learning rules derivation}
We start with KL-divergence between model and data distribution, which is equivalent to maximum likelihood:
\begin{equation}\nonumber
G = p(\pmb{v})\|p^\infty_{m}(\pmb{v}) = \sum_{2^N}\int_{(2\pi)^N} d\theta^N \, p(\pmb{v})\log(\frac{p(\pmb{v})}{p^{\infty}_{m}(\pmb{v})})
= -H(p(\pmb{v})) - \left<\log (p^{\infty}_{m}(\pmb{v}))\right>_{p(\pmb{v})}
\end{equation}
where $p(\pmb{v})$ and $p^{\infty}_{m}(\pmb{v})$ denote data and model distribution of visible units. The sum over $2^N$ denotes all $2^N$ possibilities of the modulus of visible units and the integral over $(2\pi)^N$ is the integration over the phase angles of all visible units.\\

Writing the complex weights in polar coordinates $w_{jk} = b_{jk}e^{i\theta_{jk}}$, we compute the derivative of $G$ w.r.t. $b_{jk}$:
%and $\theta_{jk}$:
\small
\begin{equation} \nonumber
\begin{split}
&\frac{\partial G}{\partial b_{jk}} 
 =-\left<\frac{\partial \log(p^{\infty}_{m}(\pmb{v}))}{b_{jk}}\right>_{p(\pmb{v})} \\
& =-\left<\frac{\sum_{2^M} \int d\theta^M \, e^{-E(\pmb{v}, \pmb{h})}(-\frac{\partial E(\pmb{v}, \pmb{h})}{\partial b_{jk}})}{\sum_{2^M} \int d\theta^M \, e^{-E(\pmb{v}, \pmb{h})}} %\ldots \right\\ 
%& 
- \left \frac{\sum_{2^{N+M}} \int d\theta^{N+M} \, e^{-E(\pmb{v}, \pmb{h})}(-\frac{\partial E(\pmb{v}, \pmb{h})}{\partial b_{jk}})}{\sum_{2^{N+M}} \int d\theta^{N+M} \, e^{-E(\pmb{v}, \pmb{h})}}\right>_{p(\pmb{v})}
\end{split}
\end{equation}
\normalsize
The sum and integral over $M$ variables denote the average over hidden unit states, each term inside becomes an average, either over the marginal distribution of the hidden variables given the visibles, or an average over the free model distribution: 
\small
\begin{equation}\nonumber
\begin{split}
\frac{\partial G}{\partial b_{jk}}  &= \left<\left<\frac{\partial E(\pmb{v}, \pmb{h})}{\partial b_{jk}}\right>_{p(\pmb{h}|\,\pmb{v})}-\left< \frac{\partial E(\pmb{v}', \pmb{h}')}{\partial b_{jk}}\right>_{p(\pmb{v}',\,\pmb{h}')}\right>_{p(\pmb{v})} \\
&=\left<|z_j||z_k|\mathrm{cos}(\theta_{jk}+\theta_k-\theta_j)\right>_{\mathrm{sample}}-\left<|z_j||z_k| \mathrm{cos}(\theta_{jk}+\theta_k-\theta_j)\right>_{\mathrm{model}}
\end{split}
\end{equation}
\normalsize
Similarly, the gradients w.r.t $\theta_{jk}$, $J_{jk}$ and $\epsilon_j$ are:
\small
\begin{equation}\nonumber
\frac{\partial G}{\partial \theta_{jk}}
= -\left<|z_j||z_k|b_{jk}\mathrm{sin}(\theta_{jk}+\theta_k-\theta_j)\right>_{\mathrm{sample}} + \left<|z_j||z_k|b_{jk}\mathrm{sin}(\theta_{jk}+\theta_k-\theta_j)\right>_{\mathrm{model}}
\end{equation}
\begin{equation}\nonumber
\frac{\partial G}{\partial J_{jk}}
= \left<|z_j||z_k|\right>_{\mathrm{sample}}-\left<|z_j||z_k|\right>_{\mathrm{model}} \qquad
\frac{\partial G}{\partial \epsilon_{j}}
= -\left<|z_j|\right>_{\mathrm{sample}}+\left<|z_j|\right>_{\mathrm{model}}
\end{equation}
\normalsize

\section{Experimental Details}
\label{Appendix: exp_detail}
\subsection{General Setup and Toy experiment}
Energy function of CAP-RBM can be written in the obvious way. In the following, $\pmb{v}$ and $\pmb{h}$ denote the complex visible and hidden units, and $\pmb{a}$ and $\pmb{b}$ the bias vectors for visible and hidden units. The energy function is then:
\begin{equation}
E = -\pmb{v}^{\dagger}\pmb{W}\pmb{h} - |\pmb{v}|^T \pmb{J} |\pmb{h}| + \pmb{a}^T |\pmb{v}| +\pmb{b}^T |\pmb{h}|
\end{equation}

Naively written in this way, this function is not necessarily real, but various simple arguments can show that we can just take the real part without causing any issue.

As in real-valued RBM, alternating parallel Gibbs sampling can be applied to hidden and visible units to sample from the model distribution. Rate, instead of sample, is used to generate an output from the CAP-RBM, for example, to compute weight updates during training, or to display visible unit activity. Rate is defined as the expected complex activity or expected modulus given fixed input to that unit. Do note that, in general, the expected modulus of a complex unit is not equal to the modulus of the expected complex activity, a slightly subtle point.

We trained our CAP-BM using 1 step Contrastive divergence (CD-1), or that followed by Persistent Contrastive divergence (PCD). The procedure of applying those methods are exactly the same as in real-valued RBMs. We observe that those methods behave as expected: CD-1 only explores the state space in vicinity to data, and forms relatively stable representations quickly. PCD learning is slower but it produces a higher quality model \cite{hinton2012practicalguidetoRBMtraining}. In our case it is able to produce a generative model for MNIST digits in CWT representation.  

We first investigate learning in the CAP-RBM on an artificial dataset of random bars with noisy phase structure. Each data sample is a 24X24 complex image, each has random numbers (2-4 each direction) of horizontal and vertical 2 pixels wide bars consist of complex numbers of modulus 1. A sinusoidal phase pattern with random overall phase offset is assigned to each bar. Additional phase perturbation is added to each pixel in a bar, sampled uniformly from $(0,0.6)$. We compare a full CAP-RBM and a CAP-RBM without $\pmb{J}$ couplings by their ability to simultaneously learn the bar-shaped amplitude pattern and the sinusoidal phase pattern. We trained both models on 40000 training examples using 10 epochs of (CD-1). The phase on bars from full CAP-RBM model appears smooth because rate, instead of samples, are shown.

\subsection{MNIST experiment}
A complex wavelet transform (CWT) was used to produce a complex representation of the original MNIST image. The CWT employs localized and oriented band pass filters with 6 orientation angles. Roughly speaking, the modulus of the resulting complex coefficients represents local power of a particular spatial frequency at different orientations, the phase represents its spatial phase value \cite{selesnick2005dualCWTreview}. We used slightly modified version of CWT (dtcwt library 0.12.0, circularly symmetric filters). The CWT were modified so that phases of filters progress mostly in the same direction when image is gradually translated. This is achieved by using complex conjugate of two of the directional filter coefficient outputs from the software package. Then average is taken over all maximum magnitudes for each frequency band across all images and the result is used to set the amplitude for each frequency band during reconstruction.

The described modified CWT was used to transform all 60000 MNIST images. Complex coefficients were normalized by the maximum modulus in its frequency band and thresholded at a cut-off modulus of 0.15 before normalizing to modulus $1$. Such thresholding is not uncommon in the CWT literature \cite{selesnick2005dualCWTreview}, and it is necessary here because coefficients with small modulus contribute little to the reconstruction but add noise to the input of other units, which deteriorates learning considerably.

Despite the thresholding of CWT coefficients, the reconstruction quality remains excellent because most of the information about digit shape is stored in phase relations between complex coefficients (Fig.~\ref{fig:main} b) second row).  

MNIST digit images have a resolution of $28 \times 28=784$ pixels. Each band pass filter in the CWT downsamples the image by a factor of 2 so there are total of 5 frequency bands. Each band has 6 directional filters. Thus, after a full DWT transform, each image is represented by a total of $(14\times 14 +7\times7+4\times4+2\times 2 + 1)\times 6= 1596$ complex numbers. In our experiments, only two bands are used for learning, resulting in $7\times7$ and $4\times4 =390$ coefficients. The highest frequency band is suppressed to limit the number of input parameters into our model, and because the high-frequency structure is relatively unimportant for expressing the relevant features of MNIST digits. Those coefficients are set to 0 during reconstruction. The low frequency bands are also suppressed in the learning because they only represent an amplitude envelope over all digits and contain little digit-specific information -- they are set to their global average during reconstruction. 

Since CAP-RBM only learns relative phase structure, visible unit activities sometimes do not have the correct global phase to yield a reasonable image reconstruction. This usually happens after large numbers of free Gibbs sampling. In cases this happened, a global phase offset was manually added based on the similarity of resulting reconstruction to hand written character. 

To produce a generative model of MNIST in CWT domain, we use a CAP-RBM with 400 hidden units and perform 10 epochs of CD1 training as initialization. Subsequently, 100 epochs of PCD training were performed without weight decay, followed by another 100 epochs of PCD training with weight decay. All experiments were implemented in numpy.

\end{appendices}

\end{document}